\documentclass[conference]{IEEEtran}
\IEEEoverridecommandlockouts
\usepackage{cite}
\usepackage{amsmath,amssymb,amsfonts}
\usepackage{algorithmic}
\usepackage{graphicx}
\usepackage{textcomp}
\usepackage{xcolor}
\def\BibTeX{{\rm B\kern-.05em{\sc i\kern-.025em b}\kern-.08em
    T\kern-.1667em\lower.7ex\hbox{E}\kern-.125emX}}

\usepackage{graphicx}
\usepackage{booktabs}
\usepackage{cite}
\usepackage{microtype}
\usepackage{xspace}
\usepackage{enumitem}
\usepackage{balance}
\usepackage{colortbl}
\usepackage{multicol}
\usepackage{multirow}
\usepackage{pifont}
\usepackage[bookmarks=false]{hyperref}
\newcommand{\name}{\textit{FedHQ}\xspace}

\renewcommand{\IEEEauthorrefmark}[1]{\textsuperscript{#1}}

\begin{document}
\title{\textit{FedHQ}: Hybrid Quantization for Federated Learning}
\author{
\IEEEauthorblockN{Zihao Zheng\IEEEauthorrefmark{1}, Ziyao Wang\IEEEauthorrefmark{2}, Jiawei Yang\IEEEauthorrefmark{1}, Hangyu Cao\IEEEauthorrefmark{1}, Zhihao Mao\IEEEauthorrefmark{1}, Maoliang Li\IEEEauthorrefmark{1}, Jiayu Chen\IEEEauthorrefmark{1}, Hailong Zou\IEEEauthorrefmark{1}, \\ Yun Liang\IEEEauthorrefmark{1}, Xiang Chen\IEEEauthorrefmark{1}$^{,  \dagger}$}
\IEEEauthorblockA{\IEEEauthorrefmark{1}Peking University \quad \IEEEauthorrefmark{2}University of Maryland, College Park \quad 
$^{\dagger}$Corresponding Author: Xiang Chen}
}

\maketitle
% ===================================================================
% sections
% ===================================================================
\begin{abstract}
Federated Learning (FL) must balance training efficiency and model accuracy across heterogeneous clients.
Post-training quantization (PTQ) avoids the extra computation of quantization-aware training (QAT), but can incur greater accuracy loss.
We propose \name, a hybrid quantization framework that assigns PTQ or QAT according to clients' hardware and data characteristics.
Hardware-based speed estimates and local data-distribution indicators guide coarse-grained global initialization, followed by fine-grained training-based adjustment of selected client strategies.
Experiments on three image-classification datasets show that \name achieves up to 2.47$\times$ training speedup over FedAQT and up to 11.15 percentage points higher accuracy than FedFQ.
\end{abstract}

\section{\textbf{Introduction}}
\label{tex:Introduction}

Federated Learning (FL) allows distributed clients to train a shared model without transmitting raw data~\cite{FedAvg,FedSGD,Fed-cbs}. Practical deployments, however, involve heterogeneous computing resources and local datasets. Differences in processing capacity and memory introduce unequal client execution times and synchronization delays~\cite{Hetero-FL-1,Hetero-FL-2, FedEL2025}, while non-IID data complicates global optimization~\cite{non-IID-1,non-IID-2, FedProx2020, FedNova2020}. Efficient FL must balance computation, communication, and accuracy across diverse clients.

\begin{figure}[!t]
    \centering
    \includegraphics[width=3.0in]{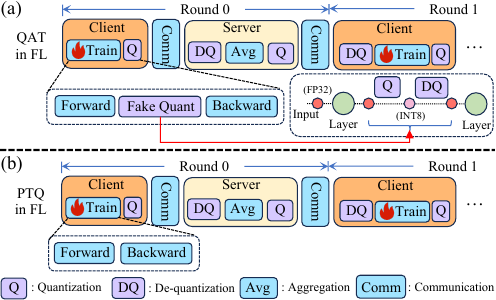}
    \vspace{-2mm}
    \caption{Quantization workflows in FL: (a) QAT; (b) PTQ.}
    \vspace{-4mm}
    \label{fig:1}
\end{figure}

Quantization reduces model representation and communication costs~\cite{DNNquant-1,AQG,FedAQ,QSGD2017,FedPAQ2020}. Two common strategies are post-training quantization (PTQ) and quantization-aware training (QAT). In the workflows of Fig.~\ref{fig:1}, PTQ performs floating-point local updates and quantizes weights for communication, avoiding additional fake-quantization operations during training. QAT instead simulates low-precision operations during local updates to mitigate quantization-induced accuracy loss, at greater computational cost~\cite{QAT-FL,EdgeQAT}. Thus, although both strategies communicate quantized parameters, they differ in local execution cost and accuracy behavior.

Existing quantized FL methods explore adaptive communication quantization~\cite{AQG, FedAQ, FedACQ}, fine-grained parameter quantization~\cite{FedFQ, FedMQPlus2025}, and quantization-aware or robust training~\cite{FedAQT, QAT-FL}.
Yet their mechanisms do not directly decide which local training strategy best suits each client. Uniform PTQ can compromise accuracy, while uniform QAT may waste computation on resource-limited clients. A natural alternative is \underline{hybrid quantization}, assigning PTQ and QAT to different clients. However, random assignments or a fixed PTQ/QAT ratio ignore clients' individual costs and potential contributions to global learning, leaving substantial room for better allocation.

Two challenges prevent straightforward allocation. 
\textbf{Challenge \ding{172}: Hardware heterogeneity} makes QAT overhead dependent on computing throughput, available memory, and feasible batch sizes~\cite{FL-survey-2,client-selection}. 
A slow QAT client can increase synchronous round latency even when other clients finish early.
\textbf{Challenge \ding{173}: Data heterogeneity} makes quantization sensitivity difficult to estimate: local training distributions vary, and the server cannot directly inspect private datasets. 
Consequently, the client that benefits most from QAT need not be the one with the greatest computational resources. 
Moreover, optimizing QAT assignments cannot be reduced to ranking clients by speed alone: fast clients may contribute little accuracy improvement, while data-sensitive clients may be slow. 
Training all assignments requires up to $2^M$ combinations for $M$ clients, making exhaustive selection impractical. 
A client-level indicator can only approximate global learning effects: the outcome of changing one strategy also depends on interactions with updates from other participating clients.

\begin{figure*}[!t]
\centering
\includegraphics[width=7in]{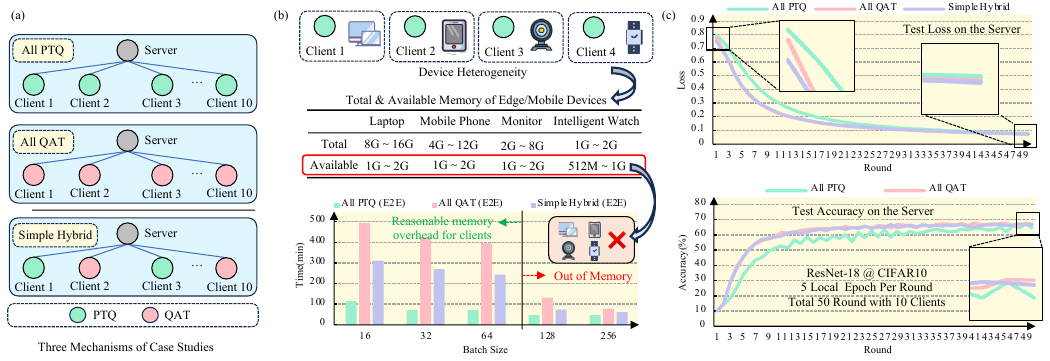}
\vspace{-2mm}
\caption{Motivating quantization study: (a) uniform and hybrid allocations; (b) resource constraints and batch-size sensitivity; (c) convergence on CIFAR-10. Device icons illustrate deployment constraints.}
\vspace{-4mm}
\label{fig:2}
\end{figure*}

We propose \name, a client-level hybrid quantization framework that integrates hardware-based speed estimates with local data-distribution indicators. 
\name first conducts \textbf{coarse-grained global initialization} to select a PTQ/QAT assignment without testing all configurations. 
It uses \textbf{fine-grained training-based refinement} to probe selected alternatives and correct potentially suboptimal decisions. 
This two-stage design seeks a favorable speed--accuracy trade-off without incurring an exhaustive search cost. 
We conduct detailed experimental evaluations to validate the effectiveness of \name.

In summary, our contributions are threefold:
\begin{itemize}[leftmargin=*]
    \item We analyze PTQ, QAT, and hybrid execution in FL, explaining why heterogeneous resources and data motivate client-specific selection.
    \item We develop complementary hardware and data-distribution indicators to guide PTQ/QAT allocation.
    \item We propose \name, combining indicator-based global initialization with selective training-based refinement.
\end{itemize}
\section{\textbf{Background}}
\label{tex:Preliminary}
In the workflows of Fig.~\ref{fig:1}, each client receives the global model, performs local updates, and uploads quantized weights~\cite{FedAQ,FedFQ}. 
The server dequantizes and aggregates these weights, then re-quantizes the global model for redistribution. 
Clients dequantize the received model before the next round.

The two strategies differ in local training. PTQ clients train in FP32 without fake quantization and quantize weights for communication. 
QAT clients additionally fake-quantize layer values to simulate lower precision, using approximate gradients through the simulated operations~\cite{FedAQT,QATFL2026}. 
These operations introduce overhead but can reduce the accuracy loss associated with quantization. 
Fake quantization itself does not imply execution with integer-only training kernels.

Therefore, hybrid allocation chooses between these local paths while retaining a common communication format.
The choice must consider resource-dependent execution cost and heterogeneous local data~\cite{FedLQ2025, LAQHC2026, HeteroFL2021, FedRolex2022}. 
A common strategy can be unsuitable for some clients; conversely, assigning QAT to more clients need not improve global accuracy monotonically. 
These considerations motivate the client-level analysis below.
% Replace sections/analysis.tex in the previous FedHQ ISCAS project.
% Figure source: original_figures/fig3.pdf (already included in that project).
% This file restores original Fig. 3; all figure numbers remain automatic.
\section{\textbf{Analysis of Quantization Strategies in FL}}
\label{Analysis}

% Declared early to keep the additional single-column figure near Section III.
% Its interpretation is given in Section III-D below.
\begin{figure}[!t]
\centering
\includegraphics[width=3.3in]{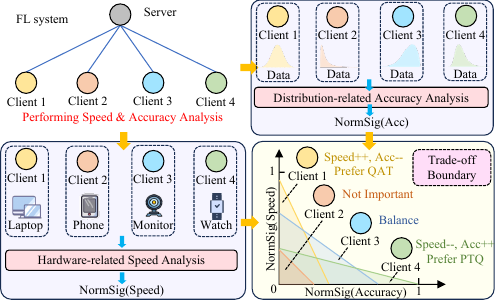}
\vspace{-2mm}
\caption{Geometric view of client-level strategy allocation. The horizontal axis represents a data-derived indicator, not measured accuracy; the small-area region is an allocation heuristic.}
\vspace{-4mm}
\label{fig:3}
\end{figure}

\subsection{\textbf{Observed Speed--Accuracy Trade-off}}
We compare All-PTQ, All-QAT, and a 1:1 hybrid allocation using ResNet-18~\cite{resnet} on CIFAR-10, with ten clients, five local epochs per round, and 50 rounds on NVIDIA A6000 GPUs (Fig.~\ref{fig:2}). Hybrid allocation yields runtime and accuracy between the two uniform strategies in this motivating case. The runtime gap decreases with increasing batch size: the relative overhead of QAT's fake-quantization operations depends on the workload, while memory constraints limit feasible batches. These observations motivate client-level allocation.

\vspace{-1mm}
\subsection{\textbf{Hardware-Related Speed Analysis}}
For client $m$, let $V_m$, $E_m$, and $B_m$ denote sample count, local epochs, and batch size. Fractions $\alpha_m,\beta_m\in(0,1]$ describe the available shares of memory $M_m$ and compute capacity $C_m$ when other workloads coexist. We estimate local training time as Eq.~\eqref{eq:time}:
\begin{equation}
T_m=\frac{V_mE_m}{B_m}\frac{P}{H_m(\alpha_mM_m,\beta_mC_m)},\quad
M_m^{B_m}\leq\alpha_mM_m,
\label{eq:time}
\end{equation}
where $P$ is the parameter count, $M_m^{B_m}$ is peak training memory, and $P/H_m$ is a hardware-profile-based batch-time estimate. The factors approximate the number of batch updates and their cost. Memory feasibility restricts $B_m$; profiles must match the model, batch size, and training path, consistent with hardware-aware optimization~\cite{HAQ2019}.
Normalize these estimates as Eq.~\eqref{eq:speed}:
\begin{equation}
s_m=\frac{T_{\max}-T_m}{T_{\max}-T_{\min}+\varepsilon},
\label{eq:speed}
\end{equation}
where extrema are taken across clients and $\varepsilon>0$ ensures stability. Larger $s_m$ indicates faster estimated execution. Synchronous round latency also depends on communication, aggregation, and the slowest participating client~\cite{TiFL2020}; $s_m$ describes local rather than end-to-end speed.

\vspace{-1mm}
\subsection{\textbf{Data-Distribution Analysis}}
Each client independently fits nine candidate families: normal, power-law, binomial, Poisson, log-normal, Student's $t$, logistic, beta, and gamma. It selects the fit with the lowest fitting error. This local preprocessing occurs once before training, and its cost is included in strategy selection. 
Let $z_{m,v}$ denote the actual feature of sample $v$ and $\mu_m$ its fitted mean. Define
\begin{equation}
r_m=V_m^{-\alpha}\sum_{v=1}^{V_m}
\operatorname{Dist}(z_{m,v},\mu_m),\quad 0<\alpha<1.
\label{eq:data}
\end{equation}
Writing $r_m=V_m^{1-\alpha}\bar d_m$ with average distance $\bar d_m$ shows that sample-count dependence is reduced, not removed. The exponent $\alpha$ differs from the resource fraction $\alpha_m$. Only summary indicators are transmitted during initialization. 

\vspace{-1mm}
\subsection{\textbf{Geometric View of the Speed--Accuracy Trade-off}}
\label{sec:geometric}
The server normalizes $r_m$ to $a_m$ and represents each client using the pair $(s_m,a_m)$. Fig.~\ref{fig:3} places $a_m$ on the horizontal axis and $s_m$ on the vertical axis. The right triangle with vertices $(0,0)$, $(a_m,0)$, and $(0,s_m)$ encodes their relative magnitudes; its area is $A_m=a_ms_m/2$.

In the illustrated heuristic, a steep triangle, such as Client~1's, favors QAT, allowing a relatively fast client to incur quantization-adaptation overhead. A shallow triangle, such as Client~4's, favors PTQ to reduce local training cost. Client~3 illustrates balanced indicators, while Client~2 occupies the small-area fallback region. These are heuristic preferences, not measured accuracy or client importance. Sec.~\ref{Design} uses the geometry for initialization and short training for refinement.

\begin{figure}[t]
\centering
\includegraphics[width=3.3in]{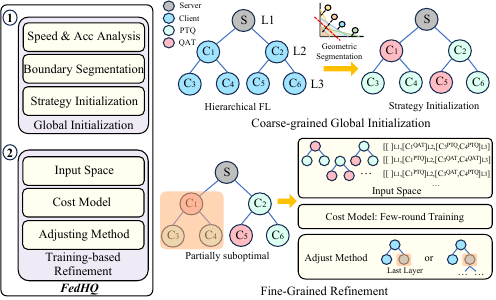}
\vspace{-2.5mm}
\caption{Overview of \name initialization and training-based refinement.}
\vspace{-4.5mm}
\label{fig:4}
\end{figure}

\vspace{-1mm}
\section{\textbf{\name: Hybrid Quantization Allocation}}
\label{Design}
Fig.~\ref{fig:4} presents the two stages of \name: global initialization assigns PTQ/QAT strategies from client indicators, while training-based adjustment evaluates selected alternatives to refine the initial allocation.

% =====================================================================
% subsection 1
% =====================================================================
\subsection{\textbf{Coarse-Grained Global Initialization}}
The upper part of Fig.~\ref{fig:4} shows indicator collection, boundary segmentation, and strategy assignment. The server normalizes client-computed hardware and data indicators and applies Fig.~\ref{fig:3}'s geometry:
\begin{equation}
q_m =
\begin{cases}
\mathrm{PTQ}, & A_m < \tau,\\
\mathrm{QAT}, & \xi s_m \geq (1-\xi)a_m,\\
\mathrm{PTQ}, & \mathrm{otherwise}.
\end{cases}
\end{equation}
The allocation boundary is determined by $\xi s_m=(1-\xi)a_m$, where $\xi=0.2$ is the default setting. Clients with $A_m<\tau$ are assigned PTQ, with $\tau=0.0625$ as an empirical threshold.
Otherwise, the weighted indicators determine the selected strategy.
This empirical fallback does not imply negligible client importance.
After local preprocessing, initialization processes a pair of scalar indicators per client, avoiding enumeration of all $2^n$ assignments. Unlike utility-guided participant selection~\cite{Oort2021}, refinement tests PTQ/QAT alternatives beyond the indicators.

\begin{figure*}[!t]
    \centering
    \vspace{-3mm}
    \includegraphics[width=6.7in]{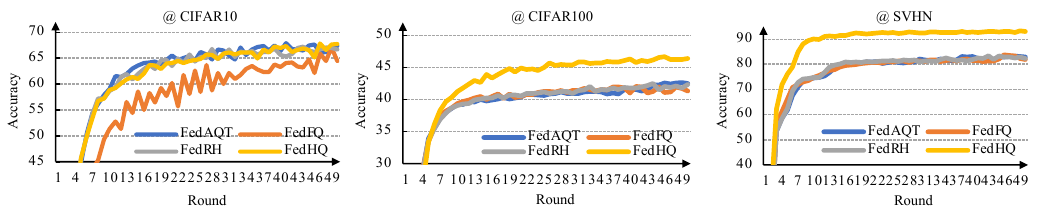}
    \vspace{-3mm}
    \caption{Accuracy Trajectories on CIFAR-10, CIFAR-100, and SVHN.}
    \vspace{-5mm}
    \label{fig:5}
\end{figure*}

% =====================================================================
% subsection 2
% =====================================================================
\subsection{\textbf{Fine-Grained Training-Based Adjustment}}
The lower part of Fig.~\ref{fig:4} links candidate construction, few-round training as the cost model, and topology-aware adjustment.

\textbf{Candidate Representation.} Refinement changes a selected subset while retaining other clients' assignments. Each candidate records both strategy choices and positions in the aggregation topology. 
For hierarchical FL~\cite{hierarchical-2}, a candidate involving $C_1$ and its children $C_3,C_4$ can be written as:
\begin{equation}
\mathcal{Q}=[\,[ \ ]_{L_1},[C_1^{\mathrm{QAT}}]_{L_2},
[C_3^{\mathrm{PTQ}},C_4^{\mathrm{PTQ}}]_{L_3}\,].
\label{eq:candidate}
\end{equation}
The subscripts identify levels, the superscripts specify strategies, and the empty entry indicates no change at that level. With $k$ binary choices, exhaustive refinement still contains $2^k$ candidates; initialization reduces the subset requiring evaluation.

\textbf{Evaluation and Refinement.} \name uses short training runs to compare alternatives involving dependent clients. 
This exposes interactions between local strategies and aggregation, at an additional probing cost. 
Early-round behavior need not preserve final-accuracy rankings. 
In the hierarchical extension, a selected leaf client's strategy is switched directly. 
For an internal client, refinement includes its dependent subtree, forms corresponding candidates, and evaluates them in a top-down order. 
Assignments outside this subtree remain fixed. 
\name evaluates selected candidates rather than exhaustive assignments, reducing search overhead.

% =====================================================================
% subsection 3
% =====================================================================
\subsection{\textbf{Execution and Cost Accounting}}
PTQ clients train in floating point before quantized communication; QAT additionally simulates quantization during updates. 
Both upload quantized parameters for aggregation, so with identical model and INT8 communication precision, assignment primarily changes local compute cost.
When initialization and refinement precede $R$ training rounds, end-to-end time is:
\begin{equation}
T_{\mathrm{end}}=T_{\mathrm{init}}+T_{\mathrm{refine}}
+\sum_{r=1}^{R}T_{\mathrm{round},r}.
\label{eq:cost}
\end{equation}
Here $T_{\mathrm{refine}}=0$ without refinement. 
Round time includes computation, synchronization, communication, and aggregation. 
Selection costs must be added only when excluded from reported training time, distinguishing the benefit of an allocation from the cost of finding it without counting probe time twice.
\section{\textbf{Experiments}}
\label{Experiment}

\vspace{-1mm}
\subsection{\textbf{Setup}}
\textbf{Settings:} We train ResNet-18 on CIFAR-10~\cite{CIFAR}, CIFAR-100~\cite{CIFAR} and SVHN~\cite{SVHN} with ten clients, five local epochs per round, and 50 communication rounds, using PyTorch, NVIDIA A6000 GPUs, and INT8 communication. We use a batch size of 128, AdamW with a learning rate of $5 \times 10^{-4}$.
Client data are partitioned using a Dirichlet distribution ($\alpha=0.5$) to simulate non-IID label skew.
\textbf{Baselines:} We compare \name with FedAQT~\cite{FedAQT}, FedFQ~\cite{FedFQ}, and random hybrid allocation (FedRH).
We measure accuracy, training speedup relative to FedAQT, and selection-inclusive time to target.

\begin{table}[!t]
\centering
\caption{Final accuracy (\%) and training speedup
relative to FedAQT.}
\vspace{-1mm}
\label{tab:main_results}
\setlength{\tabcolsep}{2.7pt}
\renewcommand{\arraystretch}{1.0}
\begin{tabular}{l|cc|cc|cc|cc}
\toprule
\multirow{2}{*}{\textbf{Task}}
& \multicolumn{2}{c|}{\textbf{FedAQT}} & \multicolumn{2}{c|}{\textbf{FedFQ}}
& \multicolumn{2}{c|}{\textbf{FedRH}} & \multicolumn{2}{c}{\cellcolor{gray!20}\textbf{\name}} \\
\cmidrule{2-9}
& \textbf{Acc.} & \textbf{Speed} & \textbf{Acc.} & \textbf{Speed} & \textbf{Acc.} & \textbf{Speed} & \cellcolor{gray!20}\textbf{Acc.} & \cellcolor{gray!20}\textbf{Speed} \\
\midrule
CIFAR-10 & 67.33 & 1.00$\times$ & 64.46 & 5.43$\times$ & 66.65 & 1.64$\times$ & \cellcolor{gray!20}\textbf{67.78} & \cellcolor{gray!20}\textbf{1.80$\times$} \\
CIFAR-100 & 42.48 & 1.00$\times$ & 41.32 & 5.28$\times$ & 42.36 & 2.09$\times$ & \cellcolor{gray!20}\textbf{46.35} & \cellcolor{gray!20}\textbf{2.47$\times$} \\
SVHN & 82.85 & 1.00$\times$ & 81.86 & 2.85$\times$ & 82.21 & 1.45$\times$ & \cellcolor{gray!20}\textbf{93.01} & \cellcolor{gray!20}\textbf{1.50$\times$} \\
\bottomrule
\end{tabular}
\vspace{-5mm}
\end{table}

\vspace{-1mm}
\subsection{\textbf{Evaluation Results}}
\textbf{Accuracy:} Fig.~\ref{fig:5} shows the accuracy trajectories. \name attains 67.78\%, 46.35\%, and 93.01\% on CIFAR-10, CIFAR-100, and SVHN, exceeding FedFQ by 3.32, 5.03, and 11.15 points and FedRH by 1.13, 3.99, and 10.80 points, respectively.
\textbf{Efficiency:} 
Tab.~\ref{tab:main_results} reports end-to-end speedups of
1.80$\times$, 2.47$\times$, and 1.50$\times$ over FedAQT, requiring 221.0, 150.7, and 163.1 minutes, respectively. 
Although FedFQ is faster, its lower accuracy confirms the need to balance training speed and quantization accuracy robustness.

\vspace{-1mm}
\subsection{\textbf{Ablation Study}}
\begin{figure}[!b]
\centering
\vspace{-4mm}
\includegraphics[width=3.0in]{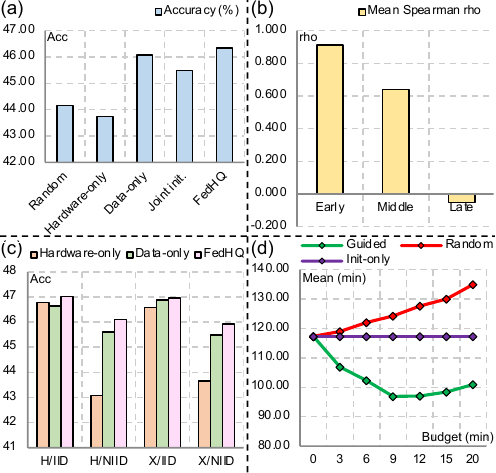}
\vspace{-2mm}
\caption{\textit{FedHQ} ablations:
(a) allocation strategies;
(b) indicator validity; 
(c) robustness under heterogeneous settings; 
(d) refinement efficiency.}
\label{fig:exp-core-draft}
\end{figure}

Fig.~\ref{fig:exp-core-draft} investigates the two-stage design.
\textbf{Allocation.}
Fig.~\ref{fig:exp-core-draft}~(a) compares five allocation policies.
Joint initialization improves upon random and hardware-only allocation, while refinement further enhances accuracy.
\name achieves the highest accuracy among the evaluated policies, demonstrating the benefit of combining hardware and data indicators with training-based refinement.
\textbf{(b) Indicator validity:} Fig.~\ref{fig:exp-core-draft}~(b) shows declining Spearman correlations between data indicators and QAT benefits, motivating training-based refinement.
\textbf{(c) Heterogeneity:} Fig.~\ref{fig:exp-core-draft}~(c) shows that \name maintains competitive accuracy under homogeneous (H)
and heterogeneous (X) hardware with IID/non-IID data.
\textbf{Refinement.}
Fig.~\ref{fig:exp-core-draft}~(d) shows that guided refinement reduces total time to 44\% accuracy.
Moderate probing budgets outperform random search and initialization only.

\subsection{\textbf{Discussion}}
\textbf{Parameter sensitivity.}
Tab.~\ref{tab:xi} shows that $\xi=0.2$ provides a favorable
accuracy--efficiency balance.
Compared with $\xi=0.3$, it improves accuracy on CIFAR-10 and SVHN with moderate additional time.
On CIFAR-100, $\xi=0.1$ further improves accuracy by 0.80 points but costs 28.6 additional minutes.
\textbf{Overhead and limitations.}
Global initialization takes 1.7 minutes, while strategy selection with refinement takes 14.2 minutes. 
Fig.~\ref{fig:exp-core-draft}~(d) includes probing overhead when measuring time to target accuracy.
Under available computational resources, we validate \name on ten-client ResNet-18 settings, with larger-scale evaluations planned for future work.

\begin{table}[!t]
\centering
\caption{Effect of $\xi$ on training time (min) / accuracy (\%).}
\vspace{-1mm}
\label{tab:xi}
\setlength{\tabcolsep}{5pt}
\renewcommand{\arraystretch}{1.0}
\begin{tabular}{lccc}
\toprule
\textbf{Dataset} & \textbf{$\xi=0.3$} & \cellcolor{gray!20}\textbf{$\xi=0.2$ (Selected)} & \textbf{$\xi=0.1$} \\
\midrule
CIFAR-10  & 210.4 / 64.31 & \cellcolor{gray!20}\textbf{221.0 / 67.78} & 240.5 / 66.42 \\
CIFAR-100 & 153.8 / 46.87 & \cellcolor{gray!20}\textbf{150.7 / 46.35} & 179.3 / 47.15 \\
SVHN      & 151.2 / 92.46 & \cellcolor{gray!20}\textbf{163.1 / 93.01} & 173.8 / 82.21 \\
\bottomrule
\end{tabular}
\vspace{-5mm}
\end{table}

\section{\textbf{Conclusion}}
\label{tex:conclusion}
We present \name, a hybrid quantization framework that combines hardware-aware and data-aware PTQ/QAT allocation with selective refinement.
Experiments demonstrate up to 2.47$\times$ training speedup over FedAQT and up to 11.15 percentage points higher accuracy than FedFQ.
Ablation studies further validate the benefits of joint allocation and refinement.

% ===================================================================
% reference
% ===================================================================
\clearpage
\balance
\bibliographystyle{IEEEtran}
\bibliography{ref/reference}

\end{document}